\documentclass[11pt,a4paper]{article}
\usepackage[utf8]{inputenc}
\usepackage{amsmath}
\usepackage{graphicx}
\usepackage{booktabs} % <-- Required for \toprule, \midrule, \bottomrule
\usepackage{float}    % <-- Required for [H] specifier
\usepackage[margin=1in]{geometry}
\usepackage{listings}  % For code
\usepackage{hyperref}  % For links
\usepackage{parskip}   % For paragraphs without indents
\usepackage{xcolor}

\usepackage{algorithm}
\usepackage{etoolbox}
\usepackage{float}
\restylefloat{algorithm}

\usepackage{algpseudocode}

\definecolor{codegreen}{rgb}{0,0.6,0}
\definecolor{codegray}{rgb}{0.5,0.5,0.5}
\definecolor{codepurple}{rgb}{0.58,0,0.82}
\definecolor{backcolour}{rgb}{0.95,0.95,0.92}

\lstdefinestyle{mystyle}{
    backgroundcolor=\color{backcolour},   
    commentstyle=\color{codegreen},
    keywordstyle=\color{magenta},
    numberstyle=\tiny\color{codegray},
    stringstyle=\color{codepurple},
    basicstyle=\ttfamily\footnotesize,
    breakatwhitespace=false,         
    breaklines=true,                 
    captionpos=b,                    
    keepspaces=true,                 
    numbers=left,                    
    numbersep=5pt,                  
    showspaces=false,                
    showstringspaces=false,
    showtabs=false,                  
    tabsize=2
}
\title{Prompt Optimization as a State-Space Search Problem}
\author{Maanas Taneja \\ University of Minnesota }
\date{\today}

\begin{document}

\maketitle

% --- Abstract (Based on your "core idea" text) ---
\begin{abstract}
Language Models are extremely susceptible to performance collapse with even
small changes to input prompt strings. Libraries such as DSpy (from Stanford
NLP) avoid this problem through demonstration-based prompt optimisation. Inspired by this, I propose an alternative approach that treats prompt optimisation as a classical state-space search problem. I model the prompt space as a graph where nodes represent prompt states and edges correspond to deliberate transformations such as shortening, adding examples, or reordering content. Using beam search and random walk algorithms, I systematically explore this space, evaluating candidates on development sets and pruning unpromising branches. Across five NLP tasks (sentiment classification, question answering, summarisation, reasoning, and natural language inference), I find that even shallow search configurations (beam width=2, depth=2) improve upon seed prompts on development sets. For instance, beam search achieves development accuracy gains from 0.40 to 0.80 on reasoning tasks, though test set improvements are more modest (0.20 to 0.50), indicating overfitting to the development heuristic. Analysis of successful optimisation paths reveals that transformations that make prompts concise appear most frequently, while verbosity operators are never selected. My results validate prompt optimization as a search problem and suggest that with greater computational resources and improved evaluation metrics, deeper exploration could yield more robust prompts that generalize beyond development sets. Code and implementation are available at [https://github.com/MaanasTaneja/PromptOptimiser].
\end{abstract}

% --- Introduction ---
\section{Introduction}

Large Language Models are remarkable mathematical constructs capable of predicting the probability of the next token in a given sequence of input tokens. 
Their ability to generate good responses therefore seems to be directly proportional to the quality of given input tokens (often generalized as a 'prompt') \cite{zhao2021}.

This also means that their performance is notoriously sensitive to the phrasing of their input.
A small, semantically-irrelevant change to a prompt can lead to drastically different, and often worse, outputs. This makes deploying LLMs in robust systems a significant challenge \cite{kojima2022}.

Recent work, such as the DSPy framework \cite{dspy}, has shown that this fragility can be mitigated by autonomously generating optimal prompts given a task. Instead of the user manually creating a good prompt through trial and error, these systems compile a declarative "signature" (e.g., text $\rightarrow$ summary) into an optimal prompt. They do this, by generating a large number of valid and quality demonstrations to teach the Large Language Model to respond in a certain way, thus optimizing the LM given a training data set of example outputs that the user is trying to reliably achieve/generate.

Inspired by this, I hypothesize that the space of all possible prompts can be modeled as a formal state-space graph, making prompt optimization a classical AI search problem.

\section{Motivation and Differences from DSPy}
\paragraph{DSPy utilises,} and in many ways exploits, the large language model’s capability to perform significantly better when provided with a handful of high-quality few-shot examples \cite{fewshot}. Rather than relying on overt, highly explicit instructions, DSPy prefers to inform the model of the type of output we wish to obtain for a given input, and then supply it with several demonstrations illustrating the desired input–output pattern. In essence, DSPy optimises a signature or task by producing the most concise possible final prompt, one that simply defines the mapping and reinforces it through well-chosen examples\cite{dspy}.

\paragraph{The DSPy optimisation process} begins with the user supplying the task signature and a small selection of seed examples. The DSPy compiler then proceeds to generate many more demonstrations similar to the seed examples using a teacher or zero shot model, bootstrapping an expanded training set  \cite{dspy}. The idea is that, by exposing the language model to a broader range of demonstrations, the module effectively “fine-tunes” its behaviour via in context learning and generalises to unseen inputs by imitating the demonstrated patterns \cite{fewshot}.

\paragraph{However,} in this setup the training data becomes the central bottleneck. With too few initial examples, the bootstrapped demonstrations risk becoming highly repetitive and stylistically close to each other, which can lead to overfitting; the model begins to answer in a mechanical, rigid manner that mirrors the demonstrations too literally. This cascading similarity in generated examples can, in some tasks, harm generalisation.

\paragraph{My hypothesis} is that directly generating and optimising the prompts themselves: by searching the prompt space and applying specific prompt-engineering transformations, may yield prompts that generalise better, akin to automated prompt construction explored in prior work \cite{shin2020autoprompt} especially in tasks requiring genuine reasoning such as NLI. This approach allows me to treat various prompt-engineering techniques as actionable “moves”, evaluate their usefulness as we traverse the prompt-search graph, and empirically compare search strategies in terms of which yield the highest gains in prompt quality.

\paragraph{Unlike DSPy,} which generates and filters demonstrations to train the model through example patterns, I instead mutate the instructions themselves. At each node, I generate a new prompt by applying one of several operator-style moves (all of which are common prompt-engineering techniques) and evaluate whether this mutation improves performance. After running this optimisation procedure, we can also extract several useful inferences beyond merely finding a stronger prompt. In particular, this framework allows us to:

\begin{itemize}
    \item \textbf{Quantify} which prompt-engineering techniques consistently improve performance,
    \item \textbf{Identify} techniques that are task-dependent rather than universally helpful,
    \item \textbf{Discover} combinations of techniques that outperform any single method in isolation,
   
\end{itemize}

To orchestrate prompt optimisation as a searching problem, I utilize these heuristics.

\begin{itemize}
    \item \textbf{States} are the textual prompts themselves.
    \item \textbf{Actions} are a set of defined "prompt transformations" or "moves" (e.g., \texttt{make\_concise}, \texttt{add\_examples}).
    \item \textbf{Heuristic} is a cost function that evaluates a prompt's performance on a small development set.
\end{itemize}

And we attempt to answer these fundamental Research Questions throughout this paper.

\begin{itemize}
    \item \textbf{RQ1:} Can classical search algorithms (e.g., Random Walk, Beam Search) meaningfully improve prompt performance, even under a highly constrained computational budget?
    \item \textbf{RQ2:} How do different search strategies (greedy vs. stochastic vs. best first) compare in their ability to explore the prompt space and find high-quality solutions?
    \item \textbf{RQ3:} Which specific prompt transformations (operators) are most frequently responsible for performance improvements?
\end{itemize}

% --- Methodology ---
\section{Methodology}
I designed and implemented a prompt optimization engine in Python. The system is composed of three core components: the state-space, the operators, and the search algorithms.

\subsection{Language Models Used in the System}

All prompt-generation, rewriting, and task inference in this work is performed using OpenAI's GPT-4o, which serves as the primary model under evaluation. Whenever a prompt $p$ is applied to an input $x$, the model prediction $f_p(x)$ is obtained via a call to GPT-4o.

For tasks that require semantic evaluation namely: \texttt{summarisation}, \texttt{nli}, and \texttt{reasoning}. I employ a more capable model, GPT-5, as a ``critic'' to judge correctness. This stronger model is never used for optimisation or rewriting prompts; it is used solely as an evaluator for the outputs generated by GPT-4o. Thus, the optimisation process searches over prompts for GPT-4o, while GPT-5 provides a higher quality analyses for complex tasks.
\subsection{Seed Prompt Generation}

Before any optimisation can begin, the system must first construct an initial
\emph{seed prompt}. This prompt is automatically generated from two inputs: a small handful of training examples, and a high level task type
(\texttt{sentiment}, \texttt{qa}, \texttt{summarisation}, \texttt{nli}, or
\texttt{reasoning}). The goal of the seed generator is to reverse engineer,
from these few examples, a reusable instruction prompt that correctly describes
the underlying task without leaking any task specific content.

Concretely, the system selects $3$--$5$ input--output pairs from the training
subset and presents them to GPT--4o, together with the declared task type. The
LLM is instructed to infer (a) what the task requires, (b) the style and
structure of the expected output, and (c) how to generalise the examples into a
concise, domain agnostic instruction. The seed generator is also instructed to never bake training examples into the seed prompt since that would defeat the purpose of our experiment and lead to extreme overfitting from step 1.

Formally, I define the seed generator as a function
\[
\text{SeedGen} : \mathcal{D}_{\text{train}} \times \mathcal{T} \rightarrow
\mathcal{P},
\]
where $\mathcal{T}$ denotes the space of task type identifiers. Given a few shot
subset $S \subset \mathcal{D}_{\text{train}}$ and the task type $t \in
\mathcal{T}$, the seed generator queries GPT--4o to produce an instruction
prompt $p_0$:
\[
p_0 = \text{SeedGen}(S, t).
\]

This $p_0$ becomes the root node of the search graph and serves as the initial
state for both Random Walk and Beam Search. All subsequent prompt variants are
mutations of this seed.

\subsection{State Representation}
A state in our search graph is represented by a \texttt{PromptNode} object, 
which contains the prompt text, a pointer to its parent node, the operator used 
to generate it, and its score (heuristic value). This structure allows us to 
reconstruct the optimization path (e.g., \texttt{seed} $\rightarrow$ 
\texttt{make\_concise} $\rightarrow$ \texttt{add\_examples}) that led to any 
given prompt.

\begin{algorithm}[!htbp]
\caption{\textsc{PromptNode} Structure}
\begin{algorithmic}[1]
\State \textbf{class} PromptNode:
\State \quad \textbf{function} Init(prompt\_text, parent = None, operator = None, score = None):
\State \qquad self.prompt\_text $\leftarrow$ prompt\_text
\State \qquad self.parent $\leftarrow$ parent
\State \qquad self.operator $\leftarrow$ operator
\State \qquad self.score $\leftarrow$ score
\State \qquad self.children $\leftarrow$ empty list
\medskip
\State \quad \textbf{function} AddChild(child):
\State \qquad append child to self.children
\medskip
\State \quad \textbf{function} GetPath():
\State \qquad path $\leftarrow$ empty list
\State \qquad node $\leftarrow$ self
\State \qquad \textbf{while} node.parent is not None:
\State \qquad \quad append node.operator to path
\State \qquad \quad node $\leftarrow$ node.parent
\State \qquad \textbf{return} reverse(path)
\end{algorithmic}
\end{algorithm}

\subsection{Evaluation Heuristic Functions}
To evaluate any prompt node, I define an evaluation heuristic
\[
\text{Eval} : \mathcal{P} \times \mathcal{D}_{\text{dev}} \rightarrow [0,1],
\]
where $\mathcal{P}$ is the space of all prompts and $\mathcal{D}_{\text{dev}}$ is the development set.
Given a prompt $p \in \mathcal{P}$, the heuristic computes the empirical accuracy using one of two
scoring functions.

\paragraph{String-Match Evaluator.}
Let $f_p(x)$ denote the model prediction under prompt $p$.  
The string-match score is:
\[
s_{\text{str}}(p,x,y) \;=\; 
\mathbf{1}\!\left[\, f_p(x) = y \,\right].
\]

\paragraph{Critic-LM Evaluator.}
Let $\mathcal{C}(f_p(x),y)$ be a binary function implemented by our more intelligent
``critic'' language model which is instructed to generate a $1$ if the prediction is judged correct
and $0$ otherwise:
\[
s_{\text{crit}}(p,x,y) \;=\; 
\mathbf{1}\!\left[\, \mathcal{C}(f_p(x),y) = \text{true} \,\right].
\]

\paragraph{Overall Heuristic.}
The evaluation heuristic averages either scoring function over the development set:
\[
\text{Eval}(p, \mathcal{D}_{\text{dev}}) = 
\frac{1}{|\mathcal{D}_{\text{dev}}|}
\sum_{(x,y)\in \mathcal{D}_{\text{dev}}}
s(p,x,y),
\]
where $s$ is either $s_{\text{str}}$ or $s_{\text{crit}}$ depending on the task being evaluated.

\subsection{Evaluator Choice by Task.}
For tasks with objective, well-defined outputs namely :\texttt{sentiment} and \texttt{qa} I use the string-matching evaluator $s_{\text{str}}$, since correctness is unambiguous and the expected outputs are short, discrete labels.  

For tasks requiring reasoning or subjective reconstruction of text \texttt{summarisation}, \texttt{nli}, and \texttt{reasoning}, the outputs cannot be reliably judged by exact string equivalence (since language model generations are not bound to discrete labels). In these cases, I use the critic-based evaluator $s_{\text{crit}}$, which employs the stronger language model to judge correctness semantically rather than syntactically.

\subsection{Critic-LM Evaluation Procedure.}
The critic is implemented as a more capable language model GPT-5 that is explicitly prompted to act as a strict yet fair evaluator. Given a prompt $p$, an input $x$, the model prediction $f_p(x)$, and the ground-truth label $y$, the critic receives all three strings and returns a binary judgement. The instruction prompt given to the critic enforces four conditions for correctness:
\begin{itemize}
    \item the model output must contain all core meaning units from the expected output,
    \item it must not introduce major unrelated content,
    \item it must not be excessively longer than the expected answer (no more than three times the length),
    \item it must match the expected output format.
\end{itemize}

The critic generates \texttt{true} if and only if all criteria are met; otherwise it generates \texttt{false}. Formally:
\[
s_{\text{crit}}(p,x,y) = 
\mathbf{1}\!\left[\mathcal{C}(f_p(x),y) = \text{true}\right],
\]
where $\mathcal{C}$ denotes the critic-LM function.

\paragraph{The heuristic score }$h(p)$ for a prompt $p$ is its average accuracy on a held-out development set.

\subsection{Prompt Operators (Moves)}
Prompt operators are text transmuting functions that transform a given input prompt, based on the definition of the operator, into a potentially more optimal prompt.

I defined a set of \texttt{GenerationMove} classes that apply transformations to a parent prompt in order to generate child prompts. These operator functions utilise specialised transformation instructions and Large Language Models to perform the prompt-mutation process.

For our experiments, we used a core set of four operators:

Formally, I define a prompt operator (or move) as a transformation function
\[
\mathcal{O} : \mathcal{P} \times \mathcal{I} \times \mathcal{D}_{\text{train}}
\;\rightarrow\; \mathcal{P},
\]
where $\mathcal{P}$ denotes the space of all prompts, $\mathcal{I}$ is the space of
transformation instructions associated with the operator, and
$\mathcal{D}_{\text{train}}$ is the training subset used for context, and as few shot example.  
Given a parent prompt $p \in \mathcal{P}$, an operator-specific instruction
$i \in \mathcal{I}$, and the training set $\mathcal{D}_{\text{train}}$, the operator
produces a new (child) prompt:
\[
p' = \mathcal{O}(p, i, \mathcal{D}_{\text{train}}).
\]

\begin{itemize}
    \item \textbf{VerboseMove:} Rewrites the prompt to be more detailed and thorough.
    \item \textbf{ShortenMove ("make\_concise"):} Rewrites the prompt to be shorter and more direct.
    \item \textbf{ReorderMove ("reorder"):} Reorganizes the content of the prompt to maximize clarity.
    \item \textbf{AddExamplesMove ("add\_examples"):} Adds 1-2 few shot examples to the prompt.
\end{itemize}

A full description of all prompt operators used in this work is provided in Appendix~\ref{appendix:operators}.

To apply any of these transformations on a given prompt input, I utilize this simple algorithm.

\begin{algorithm}[!htbp]
\caption{ApplyMove: LLM-powered prompt transformation}
\begin{algorithmic}[1]
\Function{ApplyMove}{$m, p, T$}
    \State Select few-shot examples from training set $T$
    \State Construct an instruction asking the LLM to apply move $m$ to prompt $p$
    \State Include examples from $T$ as contextual guidance
    \State $p' \gets \text{LLM}(p, m, T)$  \Comment{LLM generates rewritten prompt}
    \State \Return $p'$
\EndFunction
\end{algorithmic}
\end{algorithm}

I also devised several additional moves for this research experiment; however, I was unable to conduct experiments with them due to computational and cost constraints. These include: \texttt{ChainOfThoughtMove}, \texttt{AddConstraintsMove}, \texttt{RoleAssignmentMove}, 
and \texttt{AddDefinitionsMove}.

\subsection{Search Algorithms}

I implemented two classical search algorithms
to navigate the prompt space graph, in addition to two baselines methods of autonomous prompt optimisation.

\begin{itemize}
    \item \textbf{Seed:} Initial prompt generation by an LLM given a task signature and from a few examples from the training set.
    \item \textbf{One-Shot:} A simple approach that just asks an LLM to 
          ``improve this prompt'' in a single step. (See Algorithm 3).
    \item \textbf{Random Walk:} At each step, a random move is chosen, applied, and evaluated. 
          The best prompt seen over $N$ steps is returned. (See Algorithm 4).
    \item \textbf{Beam Search:} A best-first \cite{search} search that maintains a ``beam'' of $k$ 
          best-performing prompts at each depth level. It expands all $k$ prompts, 
          evaluates their children, and keeps only the top $k$ children for the next level. (See Algorithm 5).
\end{itemize}
\subsubsection{Search as an Optimisation Process}

I frame the goal of my search procedure as maximising the \texttt{dev\_set} evaluation score.

\paragraph{Using the search algorithms,} I begin by evaluating the seed prompt on the held-out validation set, then expand the current \texttt{PromptNode} by applying my transformation operators to generate a list of child \texttt{PromptNodes}. I then traverse the search graph in a breadth-first manner until reaching the maximum graph size, evaluating each node as it is encountered. Whenever a node achieves a higher evaluation score than the current maximum, I update the ``best prompt'' accordingly.

\paragraph{In the Random Walk search,} I choose exactly one operator at random at each node and expand only in that direction, making its performance inherently dependent on chance.

\paragraph{By contrast,} the Beam Search algorithm expands each node into four child nodes using all available moves, evaluates each child on the held-out set, and then continues expanding only the top-\(k\) nodes until the maximum graph size is reached.

\medskip

\paragraph{As a side note,} it is useful to view this optimisation procedure as an analogue to classical machine learning. The prompt text itself can be interpreted as the core mathematical object being optimised (analogous to the weights of a neural network). The search algorithm, such as \texttt{BeamSearch} or \texttt{RandomWalk}, acts as the optimiser (e.g.\ Adam or SGD) that determines which new prompt states to explore. The prompt operators, such as \texttt{make\_concise}, serve as ``update rules'' that transform the current prompt into a new one, much like a gradient update. The search is guided by the \texttt{evaluation\_metric} function, which evaluates each candidate on the \texttt{dev\_set} (our validation set) to produce a heuristic score. Finally, the best prompt discovered during search is treated as the final ``trained'' artefact and is then evaluated on the held-out test set to measure generalisation.

\begingroup
\setlength{\parskip}{0pt}      % Fix spacing above/below
\setlength{\parindent}{15pt}   % Restore normal indenting (optional)

% ---------------- ONE-HOP -------------------
\begin{algorithm}[!htbp]
\caption{One-Shot Improvement}
\begin{algorithmic}[1]
\Require Seed prompt $p$
\State Construct an ``improve this prompt'' meta-prompt using $p$
\State Query LLM to generate an improved prompt $p'$
\State \Return $p'$
\end{algorithmic}
\end{algorithm}

% ---------------- RANDOM WALK -------------------
\begin{algorithm}[!htbp]
\caption{Random Walk Search}
\begin{algorithmic}[1]
\Require Seed prompt $p$, move set $M$, train set $T$, dev set $D$, steps $N$
\State $best \gets p$
\State $current \gets p$
\For{$t = 1$ to $N$}
    \State Choose random move $m \in M$
    \State $p' \gets m.\text{apply}(current, T)$ \Comment{Use $T$ as few-shot context}
    \State $score \gets \text{Evaluate}(p', D)$
    \If{$score > \text{Evaluate}(best, D)$}
        \State $best \gets p'$
    \EndIf
    \State $current \gets p'$ \Comment{Random walk steps forward}
\EndFor
\State \Return $best$
\end{algorithmic}
\end{algorithm}

% ---------------- BEAM SEARCH -------------------
\begin{algorithm}[!htbp]
\caption{Beam Search over Prompt Space}
\begin{algorithmic}[1]
\Require Seed prompt $p$, move set $M$, train set $T$, dev set $D$, beam width $k$, depth $d$
\State $root \gets p$
\State Evaluate $root$ on $D$
\State $beam \gets \{root\}$
\State $best \gets root$

\For{$level = 1$ to $d$}
    \State $candidates \gets \emptyset$
    \For{\textbf{each} node $n$ \textbf{in} $beam$}
        \For{\textbf{each} move $m$ \textbf{in} $M$}
            \State $p' \gets m.\text{apply}(n, T)$ \Comment{Prompt rewritten using $T$}
            \State $score \gets \text{Evaluate}(p', D)$
            \State Add $(p', score)$ to $candidates$
            \If{$score > \text{Evaluate}(best, D)$}
                \State $best \gets p'$
            \EndIf
        \EndFor
    \EndFor
    \State Sort $candidates$ by descending score
    \State $beam \gets$ top $k$ prompts from $candidates$
\EndFor

\State \Return $best$
\end{algorithmic}
\end{algorithm}

\endgroup

% --- Experimental Setup ---
\section{Experimental Setup}
To evaluate and run experiments with this proposed system, I created datasets utilizing GPT-5 \cite{gpt5} across five commonly seen  NLP tasks.

\begin{itemize}
    \item \textbf{Sentiment Classification:} Classify text as positive, negative, or neutral.
    \item \textbf{Question Answering (QA):} Provide a short, factual answer to a question.
    \item \textbf{Summarization:} Condense a statement into a subject-verb-object phrase.
    \item \textbf{Complex Reasoning:} Provide reasoning for a particular assertion.
    \item \textbf{Natural Language Inference (NLI):} Classify premise/hypothesis pairs as entailment, and contradiction.
    
\end{itemize}
For each task, the dataset was split 25\% for training (used for seed generation and by the prompt operators to generate an updated prompt), 25\% for development (the heuristic), and 50\% for testing (final generalization score).

Due to numerous computational and cost constraints, I used a shallow search setting: the Random Walk was limited to $N = 5$ steps, and the Beam Search was restricted to a beam width of $k = 2$ and a depth of $d = 2$.

\subsection{Dataset Construction and Limitations}
All datasets used in this work were synthetically generated using GPT-5 to enable rapid experimentation and controlled evaluation across diverse task types. Each task dataset contains 20 examples total: 5 training examples (25\%, used for seed generation and operator context), 5 development examples (25\%, used as the optimization heuristic), and 10 test examples (50\%, held out for final evaluation).

While synthetic data enables systematic exploration of the prompt optimization framework, it introduces several limitations. Generated examples may lack the linguistic diversity, edge cases, and real-world complexity present in established benchmarks such as GLUE, SuperGLUE, or Big-Bench. Additionally, the small dataset size (particularly the 5-example development set) may amplify overfitting and increase variance in heuristic scores. Future work should validate these findings on larger, naturally occurring datasets to assess whether the observed optimization patterns generalize to production settings.

% --- Results ---
\section{Results}
The following sections present the findings for my three research questions, based on the data from the experimental runs.

\subsection{RQ1: Do search algorithms improve prompts?}
Primary data collection and investigation were concentrated on answering whether classical search algorithms can meaningfully improve prompts. As shown in Table \ref{tab:dev_results} and Table \ref{tab:test_results}, I observed that:
\paragraph{Both} random walk and beam search consistently discover prompts that outperform the original seed on the dev set (e.g., in reasoning, beam search improves the dev score from 0.40 to 0.80). Beam search reliably finds the strongest dev set prompts, as expected for a best-first strategy. Random walk also finds improved prompts; however, I do not interpret this as evidence of an effective exploration strategy. Since, its apparent success is a consequence of the particular operator set used in these experiments: all four operators tend to strictly benefit prompts (the strategies these operators employ are rarely known to cause a decrease in performance or generalization), across most tasks. In this scenario, random traversal is likely to land on a better prompt. The one-shot greedy baseline rarely beats the seed, suggesting that prompt optimisation generally requires multi-step transformations of the seed prompt.

The erratic behaviour of random walk across tasks further supports this interpretation. In more subjective and complex tasks that require the language model to balance multiple factors (brevity, tone, style, and task-specific content, as in summarisation), especially when deterministic string-matching metrics cannot reasonably evaluate correctness (necessitating the use of a \texttt{CriticLM}), random walk is not universally helpful. In these settings it fails to improve upon the seed prompt, whereas beam search continues to make progress. Conversely, in tasks where all operators are consistently helpful (sentiment, QA, NLI, reasoning), random walk performs surprisingly well, simply because any randomly chosen move is likely to be beneficial.

\paragraph{An important phenomenon} emerges when comparing dev set and test set performance (Table~\ref{tab:test_results}). While dev set scores often improve substantially, test set scores  plateau. For example, in reasoning, the dev score improves from 0.40 to 0.80, yet the test score only moves from 0.20 to 0.50. This clearly speaks to classical overfitting: beam search aggressively exploits the dev set evaluation metric to achieve better scores each layer traversal, hindering its ability to generalize outside of our validation set.

\paragraph{RQ1 Results Tabulated} Even with shallow search, state-space exploration improves prompts, but "better dev prompt" $\neq$ "better generalizing prompt," remains to be an issue which I speak more to in RQ2.

% --- The dev set table ---
\begin{table}[H]
\centering
\caption{Dev set accuracy/score (heuristic performance) across tasks and optimization methods.}
\label{tab:dev_results}
\begin{tabular}{lrrrr}
\toprule
method &  seed &  one\_shot &  random\_walk &  beam\_search \\
task          &       &          &              &              \\
\midrule
nli           &  1.00 &     1.00 &         1.00 &         1.00 \\
qa            &  0.60 &     0.60 &         0.80 &         0.80 \\
reasoning     &  0.40 &     0.20 &         0.60 &         0.80 \\
sentiment     &  1.00 &     1.00 &         1.00 &         1.00 \\
summarization &  0.40 &     0.40 &         0.40 &         0.60 \\
\bottomrule
\end{tabular}
\end{table}

% --- The test set table ---
\begin{table}[H]
\centering
\caption{Test set accuracy/score (generalization) across tasks and optimization methods.}
\label{tab:test_results}
\begin{tabular}{lrrrr}
\toprule
method &  seed &  one\_shot &  random\_walk &  beam\_search \\
task          &       &          &              &              \\
\midrule
nli           &  1.00 &     1.00 &         1.00 &         1.00 \\
qa            &  1.00 &     1.00 &         1.00 &         1.00 \\
reasoning     &  0.20 &     0.30 &         0.50 &         0.50 \\
sentiment     &  1.00 &     0.90 &         1.00 &         1.00 \\
summarization &  0.30 &     0.20 &         0.30 &         0.30 \\
\bottomrule
\end{tabular}
\end{table}

\subsection{RQ2: How do different search strategies compare?}
To understand how search structure affects optimization, let us first compare the behavior of each algorithm.

\subsection{Search Behavior}
\paragraph{One-shot} attempts to improve the prompt by providing a language model with a general instruction to do so given current prompt (in essence only explores one more node in our prompt space)

\paragraph{Random Walk}tries to explore the space more thoroughly but relies on random chance rather than heuristics to explore more optimal routes; thus finding occasional improvements unstably.

\paragraph{Beam Search} (even at 2x2) explores the graph more thoroughly and attempts to bias its search in the most promising directions utilizing best first strategy, consistently finding the best dev scores.

\paragraph{In conclusion,} Beam search consistently outperforms random walk on the dev set (Table \ref{tab:dev_results}). Even with width=2, depth=2, it expands a structured frontier and quickly finds  optimizations. One-Shot almost never works, confirming that meaningful prompt optimization requires multi-step compositions of transformations.

\paragraph{Why Test Scores Flatten}  
There are a few reasons why strong dev set gains do not always carry over to the test set in my experiments.

\begin{enumerate}
    \item \textbf{Overfitting to the Dev Set.}  
    Some operators, especially \texttt{AddExamples} can unintentionally bake in examples or patterns from the training or dev split directly into the prompt. Beam search then keeps selecting those prompts simply because they score highly on the dev examples, even if those improvements are extremely specific to that split. Over time the search essentially "overfits" the dev set heuristic, producing prompts that perform very well  on the dev set but do not generalise.

    \item \textbf{Metric and Evaluator Misalignment.}  
    For tasks with objective, clean outputs (sentiment, QA, NLI), string matching works very well, and we quickly maximise the test scores. However, for more complex and subjective tasks such as reasoning and summarisation, I rely on a \texttt{CriticLM} to judge correctness. Since the critic is a language model itself, it is sometimes overly rigid or stylistically sensitive: it may mark an answer wrong simply because the wording or phrasing differs from the ground truth, even when the underlying reasoning or summary is technically accurate. This introduces noise, and the search optimises to please the \texttt{CriticLM} than for actual semantic correctness. A stronger critic prompt or a better evaluation metric would almost certainly help.

    \item \textbf{Shallow Search.}  
    Finally, since my search setting is intentionally shallow (beam width $k=2$, depth $d=2$, or a random walk of only $N=5$ steps) due to compute and cost constraints. This limits how much of the prompt space the algorithms can realistically explore. It is quite possible that deeper or wider searches would yield prompts that generalise better.
\end{enumerate}

\paragraph{In conclusion,} Beam search is the most reliable strategy under compute constraints. However, prompt optimization is only as good as its reward signal, and noisy evaluators flatten test performance.

\subsection{RQ3: Which transformations contribute most?}
To investigate which operations were most useful, I analyzed the transformation paths that led to our final optimized prompt artifact. Further, I computed the frequency of each operator across all successful paths found by Random Walk and Beam Search.

% --- The RQ3 results table ---
\begin{table}[H]
\centering
\caption{Frequency of operators in successful optimization paths across all tasks.}
\label{tab:rq3_freq}
\begin{tabular}{lr}
\toprule
Operator (Move) & Frequency \\
\midrule
make\_concise   & 4         \\
add\_examples   & 2         \\
reorder         & 2         \\
make\_verbose   & 0         \\
\bottomrule
\end{tabular}
\end{table}

\paragraph{The results,} shown in Table \ref{tab:rq3_freq}, are unambiguous, and lead to three major conclusions.
\begin{enumerate}
    \item \textbf{Conciseness is key:} 
    The \texttt{make\_concise} operator was the most frequently used. 
    This confirms that models respond better to short, unambiguous instructions 
    and that overly verbose prompts can weaken task alignment. Concise prompts also may lead to better generalization, since the probability chain of longer token sequences can be easily skewed in various specific directions in our output space.

    \item \textbf{Examples are useful:} 
    The \texttt{add\_examples} operator was also consistently used. 
    This suggests LLMs benefit from clear input/output pairs that anchor 
    the instruction and prevent structural drift.

    \item \textbf{Verbosity rarely helps:} 
    The \texttt{make\_verbose} operator did not appear in any successful optimization path, 
    suggesting it may harm performance by diluting the task structure.
\end{enumerate}

\paragraph{In conclusion, }
the search process implicitly learns which operators are stable and reward-aligned. Conciseness and examples consistently form the backbone of optimized prompts in the experiments.

% --- Discussion & Limitations (From your draft) ---
% --- Discussion & Limitations (From your draft) ---
\section{Discussion \& Limitations}
\subsection{Discussion}
This research proposal treated prompt optimization as a search problem. And I was left with three important results.
\begin{itemize}
    \item \textbf{Validation of Searching Prompt-Space:}
    These findings can serve as a strong proof of concept for framing prompt optimization as a formal search problem. Despite using extremely shallow search depths (D=2) and a narrow beam width (W=2) due to computational and cost constraints, the methods employed consistently produced prompts that improved upon the seed generation. This validates the core hypothesis that the prompt landscape can be searched for more optimized nodes using classical search techniques. It stands to reason that with a larger compute budget, running deeper (e.g., \( D > 5 \)) and broader (e.g., \( W > 5 \)) searches would allow the optimizer to escape local optima and discover significantly more performant and robust prompts.

    \item \textbf{Certain Prompt Engineering methods outperform others:} 
    RQ3 analysis revealed \texttt{make\_concise} and \texttt{add\_examples} 
    appear most frequently in successful paths. This aligns with the consensus that 
    shorter, cleaner instructions improve reliability.

    \item \textbf{Beam Search $>$ Random Walk:}
    Beam Search yielded more stable improvements on dev sets, but test-set performance 
    often ``flattened,'' especially when using the Critic-LM evaluator. This suggests 
    overfitting concerns using small data sets, and that the Critic-LM is sometimes an overly harsh  evaluator, and can be improved upon.
\end{itemize}
\subsection{Limitations and Future Directions}
Although this project demonstrates the viability of search-based prompt optimization, several important limitations must be addressed in future work:

\paragraph{Computational Constraints.} Time and cost constraints prevented deeper exploration of the prompt space. Our beam width (W=2) and depth (D=2) represent minimal proof-of-concept settings. Methods like OPRO \cite{opro} typically use 100--200 optimization steps; our equivalent is approximately 8 prompt evaluations per task. Future work should explore W$\geq$5, D$\geq$5 to determine whether deeper search consistently improves generalization or exacerbates overfitting.

\paragraph{Small Synthetic Datasets.} Using only 5 development examples per task likely amplifies both overfitting and evaluation noise. The synthetic nature of the data may not capture the full complexity of real-world benchmarks. Validation on established datasets (GSM8K, MMLU, Big-Bench Hard, SuperGLUE) with larger development sets (100+ examples) is essential to assess whether these optimization patterns hold in production settings.

\paragraph{Limited Operator Set.} I tested only 4 operators due to cost constraints. Future work should explore richer transformation sets including \texttt{ChainOfThoughtMove}, \texttt{RoleAssignment}, \texttt{AddDefinitions}, and \texttt{AddConstraints}. Additionally, operators could be learned from data rather than manually designed, or proposed by LLMs themselves.

\paragraph{Evaluation Metrics.} String-matching is too strict for open-ended tasks, while Critic-LM evaluation introduces bias and noise. The critic may penalize stylistic variations that preserve semantic correctness. Better evaluation strategies include: (1) semantic similarity metrics (BERTScore, embedding distance), (2) human evaluation on sampled outputs, (3) task-specific metrics (F1 for NLI, ROUGE for summarization), or (4) reward models trained on human preference data.

\paragraph{Overfitting Mitigation.} The substantial dev-test performance gap (e.g., reasoning: 0.80 dev, 0.50 test) suggests search is exploiting idiosyncrasies of the small development set. Future work should explore: (1) regularization techniques (penalizing prompt length, limiting examples), (2) k-fold cross-validation during search, (3) ensemble methods combining multiple optimization runs, and (4) early stopping criteria based on held-out validation subsets.

\paragraph{Strong Seed Prompts.} For some tasks (NLI, Sentiment), automatically generated seed prompts achieved 100\% accuracy, creating a ceiling effect where optimization cannot demonstrate improvement. Tasks should be selected to allow meaningful optimization headroom.

\paragraph{Cost and Latency.} Each task run required 30--45 minutes and several dollars in API costs, dominated by GPT-5 as CriticLM. Future work could reduce costs by: (1) using smaller, faster models as critics, (2) leveraging open-source local models (Llama, Qwen, DeepSeek), (3) caching evaluations for identical prompts, or (4) using learned reward models instead of LLM-as-judge.
\section{Related Work}
\subsection{Prompt Optimization Approaches}
Several recent methods have explored automated prompt optimization, each taking different approaches to the search problem.

\paragraph{OPRO \cite{opro2023} (Optimization by Prompting)} treats prompt refinement as a meta-optimization task. The LLM is provided with examples of prompts paired with their performance and is asked to generate improved variants. This approach relies primarily on the generative capabilities of the model rather than an explicit search structure.

\paragraph{DSPy} focuses on optimizing demonstrations and few-shot examples rather than instruction text itself. Its declarative pipeline compiles into self-improving procedures that learn how to construct better exemplars. This work is complementary, as we focus on transformations of the instruction rather than demonstration synthesis.

\paragraph{APE \cite{ape}} frames instruction generation as natural-language program synthesis. It samples large candidate sets using forward or reverse generation templates and scores them using metrics such as execution accuracy. APE performs a broad sampling-and-filtering process rather than a structured local search.

\subsection{Comparison to Our Approach}
Table~\ref{tab:method_comparison} highlights the conceptual distinctions among these approaches.

\begin{table}[H]
\centering
\caption{High-level comparison of prompt optimization methods.}
\label{tab:method_comparison}
\begin{tabular}{p{2.5cm}p{4.2cm}p{3.0cm}}
\toprule
\textbf{Method} & \textbf{Core Mechanism} & \textbf{Primary Focus} \\
\midrule
DSPy & Generates and refines demonstrations & Few-shot exemplars \\
OPRO & LLM-driven meta-optimization of prompts & Direct prompt rewriting \\
APE & Sample large candidate sets; score-and-filter & Instruction induction \\
Ours & Local state-space search (beam, random walk) & Instruction-level refinement \\
\bottomrule
\end{tabular}
\end{table}

This work differs by: (1) explicitly modeling prompt optimization as a state-space graph with defined rewrite operators, (2) using classical search algorithms rather than broad sampling or meta-prompting, and (3) analyzing the contribution of individual transformation operators. While the search depth is deliberately shallow for feasibility, these results demonstrate that structured search alone can systematically improve prompts.

\section*{Acknowledgments}
This work was completed as an independent research project for CSCI 4511W by Professor James Moen at the University of Minnesota. Computational resources were limited by OpenAI API costs; deeper exploration with larger datasets and search configurations remains for future work.

% --- References (Placeholder) ---

\clearpage

\section*{Appendix A: Prompt Operator Definitions}
\label{appendix:operators}

\begin{table}[h]
\centering
\small
\begin{tabular}{p{3cm}p{6.5cm}p{5cm}}
\toprule
\textbf{Move} & \textbf{Description} & \textbf{Example Transformation} \\
\midrule

\texttt{make\_verbose} 
& Expands the prompt with additional detail, clarification, and explicit guidance while preserving the required output format.
& ``Classify the sentiment.'' $\rightarrow$ ``Analyse the input carefully and provide a detailed sentiment judgement.'' \\[0.35cm]

\texttt{add\_cot} 
& Adds chain of thought instructions encouraging step by step reasoning. \cite{cot}
& ``Is the answer correct?'' $\rightarrow$ ``Break the problem into steps and think through the reasoning before giving the final answer.'' \\[0.35cm]

\texttt{make\_concise} 
& Makes the prompt shorter and more direct without changing the task.
& ``Carefully read the input and provide a highly accurate classification.'' $\rightarrow$ ``Read the input and classify it accurately.'' \\[0.35cm]

\texttt{reorder}
& Reorganises the structure of the prompt for better clarity and flow.
& Moves ``Output Format'' before ``Task Description'' $\rightarrow$ ``Task $\rightarrow$ Requirements $\rightarrow$ Output''. \\[0.35cm]

\texttt{add\_examples}
& Adds 1--2 few shot examples showing the expected input/output format.
& Appends: \textit{Input: X Output: Y}. \\[0.35cm]

\texttt{add\_constraints}
& Adds reliability constraints such as avoiding hallucination, adhering to input information, staying concise, and following format.
& Adds rules like: ``Do not hallucinate facts; use only information from the input.'' \\[0.35cm]

\texttt{add\_role}
& Prepends a strong persona to shape model behaviour.
& ``You are a highly precise and detail-oriented analyst.'' \\[0.35cm]

\texttt{add\_definitions}
& Adds definitions for labels, categories, or output types to reduce ambiguity.
& ``Entailment: hypothesis logically follows from premise.'' \\[0.35cm]

\bottomrule
\end{tabular}
\caption{Full list of prompt transformation operators used in the search procedure.}
\label{tab:operators_full}
\end{table}

\end{document}